\documentclass[runningheads]{llncs}
\usepackage{makeidx}
\usepackage{graphicx}
\usepackage{amsmath,amssymb} 
\usepackage{color}
\usepackage{multirow}

\makeatletter
\newcommand{\printfnsymbol}[1]{%
  \textsuperscript{\@fnsymbol{#1}}%
}
\makeatother

\begin{document}
	\pagestyle{headings}
	\mainmatter

	\def\GCPR19SubNumber{69}

	\title{Points2Pix: 3D Point-Cloud to Image Translation using conditional GANs}

	
	\titlerunning{Points2Pix: 3D Point-Cloud to Image Translation using conditional GANs}
	\authorrunning{Milz et al.}
\author{
Stefan Milz\inst{1} \thanks{Equal contribution} \and
Martin Simon\inst{1} \printfnsymbol{1} \and
Kai Fischer\inst{1} \printfnsymbol{1} \and
Maximillian P\"opperl \inst{1}  \and \\
Horst-Michael Gross \inst{2} 
}

\institute{
$^1$ Valeo Schalter und Sensoren GmbH, Germany \\
$^2$ Ilmenau University of Technology, Germany 
}

	\maketitle

\begin{abstract}
We present the first approach for 3D point-cloud to image translation based on conditional Generative Adversarial Networks (cGAN). The model handles multi-modal information sources from different domains, i.e. raw point-sets and images. The generator is capable of processing three conditions, whereas the point-cloud is encoded as raw point-set and camera projection. An image background patch is used as constraint to bias environmental texturing. A global approximation function within the generator is directly applied on the point-cloud (Point-Net). Hence, the representative learning model incorporates global 3D characteristics directly at the latent feature space. Conditions are used to bias the background and the viewpoint of the generated image. This opens up new ways in augmenting or texturing 3D data to aim the generation of fully individual images. We successfully evaluated our method on the KITTI and SunRGBD dataset with an outstanding object detection inception score.
\end{abstract}

\section{Introduction}

Domain translation is a well known and widely applied problem. It is typically treated in computer graphics or computer vision. Most research focuses on image-to-image translation \cite{DBLP:journals/corr/IsolaZZE16,DBLP:journals/corr/ZhuPIE17,wang2018pix2pixHD}. Examples are Semantic-Labels to Image (e.g. Labels to Street-Scene, Labels to Facades ) or Image conversions (e.g. Day to Night, Black-and-White to Color). Those techniques deal with real domain translation problems, since they convert semantic sensor-independent context into realistic RGB image data or vice versa. However, domain translation is performed on top of images. Both domains encode the information as RGB values in pictures with a spatial dependency. We call that single mode domain translation:  
\begin{equation}
\begin{aligned}
G_{x \rightarrow y}: x \rightarrow y &\;\;\;\; x,y \in \mathbb{R}^{w\times h \times 3}\\
G_{y \rightarrow x}: y \rightarrow x &
\end{aligned}
\end{equation}
Whereas, $G_{x \rightarrow y}, G_{y \rightarrow x}$ describe the translation functions between both image domains $x$, $y$ with fixed image sizes: $h$ (height), $w$ (width). 

We propose a novel multi-modal domain translation model using the example of 3D point-cloud to image translation. The treated problem is formally known as:
\begin{align}
G_{p \rightarrow y}: p \rightarrow y &\;\;\;\; p \in \mathbb{R}^{n\times  3}\;\; y \in \mathbb{R}^{w\times h \times 3}
\end{align}
Here, n describes the number of points within the point-set. Our work is limited to $G_{p \rightarrow y}$ (not $G_{y \rightarrow p}$). Therefore an extensive new architecture is presented as combination of a typical encoder-decoder for image segmentation (UNet: \cite{DBLP:journals/corr/RonnebergerFB15}) as proposed by \cite{DBLP:journals/corr/IsolaZZE16}. More important is the models second input, where the architecture incorporates the real point-set to add 3D characteristics into the global feature space for constraint based individual image generation. We put conditions as viewpoint dependent projection and background image patches for fully individual image generation in compliance with 3D specifications (conditions: background, shape, distance, viewpoint). 

\section{Related Work}
\subsection{Image generation}
\subsubsection{Handcrafted Losses}
As image generation could be reduced to per-pixel classification/regression with a wide application area it turns out to have a long tradition \cite{DBLP:journals/pami/ShelhamerLD17,DBLP:journals/corr/XieT15,DBLP:journals/corr/YooKPPK16,IizukaSIGGRAPH2016}. Those applications suppose a conditionally unstructured loss applied on the output space, i.e. a pixel independence in terms of semantic relationship is supposed. The performance of those approaches strongly depends on the loss design, e.g. semantic segmentation \cite{DBLP:journals/corr/PaszkeCKC16}.

\subsubsection{Conditional GANs}
Conditional GANs (cGAN) instead learn structured losses that affect the overall output in form of a joint improvement \cite{DBLP:journals/corr/IsolaZZE16}. In common, the cGAN is applied in a conditional setting. For image generation researchers were setting variable conditions: e.g. discrete labels \cite{DBLP:journals/corr/MirzaO14,DBLP:journals/corr/DentonCSF15}, text \cite{DBLP:journals/corr/ReedAYLSL16} and images \cite{DBLP:journals/corr/IsolaZZE16,DBLP:journals/corr/ZhuPIE17,wang2018pix2pixHD}. 

In general the cGAN performs a mapping function $G$, called generator, based on a condition $c$ and a random noise vector $z$ to generate an image $y$: 
\begin{equation}
G: \{c,z\} \rightarrow y \;\;\;\; y \in \mathbb{R}^{w\times h \times 3} \;\;\;\;
c = \left\{
\begin{array}{ll}
\in \mathbb{R} & \rightarrow\textrm{label-to-image} \\
\in \mathbb{R}^{t} & \rightarrow\textrm{text-to-image} \\
\in \mathbb{R}^{w\times h \times 3} &\rightarrow\textrm{image-to-image} \\
\end{array}
\right.
\label{equ:noise}
\end{equation}
For image-to-image translation \cite{DBLP:journals/corr/IsolaZZE16} proposes a \textit{U-Net} like structure for $G$. To create realistic images at higher resolutions (e.g. 1024 x 2048) \cite{wang2018pix2pixHD} recommend a pyramidal approach for $G$ similar to a \textit{PSPNet} \cite{DBLP:journals/corr/ZhaoSQWJ16}.

In general, the cGAN is composed by $G$ and a competing discriminator $D$, which distinguishes between real images and created fake ones. A well-established discriminator network is the \textit{Patchgan} \cite{DBLP:journals/corr/Li16p} proposed by \cite{DBLP:journals/corr/IsolaZZE16}. Derived from that the competing objective of the cGAN could be described by its loss $L_{\textrm{cGAN}}$:
\begin{equation}
L_{cGAN}(G, D) = \mathbb{E}_{c,y}\{\log(D(c, y))\} +  \mathbb{E}_{c,z}\{\log(1-D(c,G(c,z))\}
\label{equ:loss}
\end{equation}

\subsection{Point-cloud processing}

High requirements for perception tasks of robotic applications enforced the usage of 3D sensors, e.g. RGBD-cameras \cite{DBLP:journals/corr/SongX15},  Lidar (Valeo SCALA). Research progress in the field of 3D point-cloud processing received a boost in the recent years. 
In principle, point-clouds have specific properties that clearly distinguish them from images. Hence, specific processing models are needed. Points usually are not ordered, there is no grid that encodes the 3D position as an image does. The overall category of a point-set is influenced by the interaction of points among others. Only the global sum of the points forms a shape with a meaning. Last, point-sets are invariant to basic transformations like translation or rotation. Therefore the combination of 3D points clouds and machine learning is indispensable. The processing type could be categorized into the following three classes.

\subsubsection{Real 3D Point-cloud processing}
\cite{qi2016pointnet} proposed the first neural network architecture \textit{Point-Net} that handles natural points sets for classification and segmentation tasks with outperforming segmentation results on ShapeNet \cite{DBLP:journals/corr/ChangFGHHLSSSSX15}: \textbf{mIoU 83.7}. The model does not use convolutional layers, but fully connected ones and directly processes the coordinates of the point-set (size $n$): $p = {x_1 ..., x_n}$  with $p \in \mathbb{R}^{n\times 3}$. A chain of local transformations $h$ on the point-set followed by a global max-pooling layer is used to create an overall feature space, i.e. a global approximation function:
\begin{equation}
\begin{aligned}
f(x_1, ... ,x_n) \approx g(h(x_1), ..., h(x_n)) \;\;\;\ \;\;\;\
f:& 2^{\mathbb{R}^{n\times3}}\rightarrow \mathbb{R} \\
h:& \mathbb{R}^{n\times3}\rightarrow \mathbb{R}^K \\
g:& \{\mathbb{R}_1^K \times ... \times \mathbb{R}_n^K\} \rightarrow \mathbb{R}
\end{aligned}
\label{equ:point}
\end{equation}
I.e. The overall meaning $f$ (e.g. object class) of a point set $p$ is approximated by $g$. The advantage of the architecture is that it is robust against unordered point-clouds and transformations. The independence form the viewpoint variance helps to train with less training samples. Due to disadvantage for learning global features of large point sets the authors developed a second version \textit{Point-Net++} \cite{DBLP:journals/corr/QiYSG17}.

\subsubsection{Voxelization}
Voxelization approaches make use of the findings performing CNNs on images. Therefore, 3D data is converted to voxels or grid cells. After pre-processing standard machine learning architectures are applied. Unordered point-sets are avoided. Famous applications are 3D object detectors like \cite{DBLP:journals/corr/abs-1803-06199,DBLP:journals/corr/ChenMWLX16,Luo_2018_CVPR,ku2018joint}

\subsubsection{Combined models}
Combined models have often shown most robust results (e.g. 3D object detection) and mostly make use of different sensor types. \cite{DBLP:journals/corr/abs-1711-06396} investigated a method based on many local \textit{Point-Nets} followed by a global 3D CNN. \cite{DBLP:journals/corr/abs-1711-08488} architectures works the other way around. With the aid of a camera frustum points are filtered using a camera object detector. The filtered points are processed for 3D object detection with only one \textit{Point-Net} \cite{qi2016pointnet} up to the last global max pooling layer ending in a $1\times 1024$ general feature space. A 8 bit depth projection using the given camera projection matrix ($c_2$)

\begin{figure*}[!t]
\includegraphics[width=1.0\textwidth]{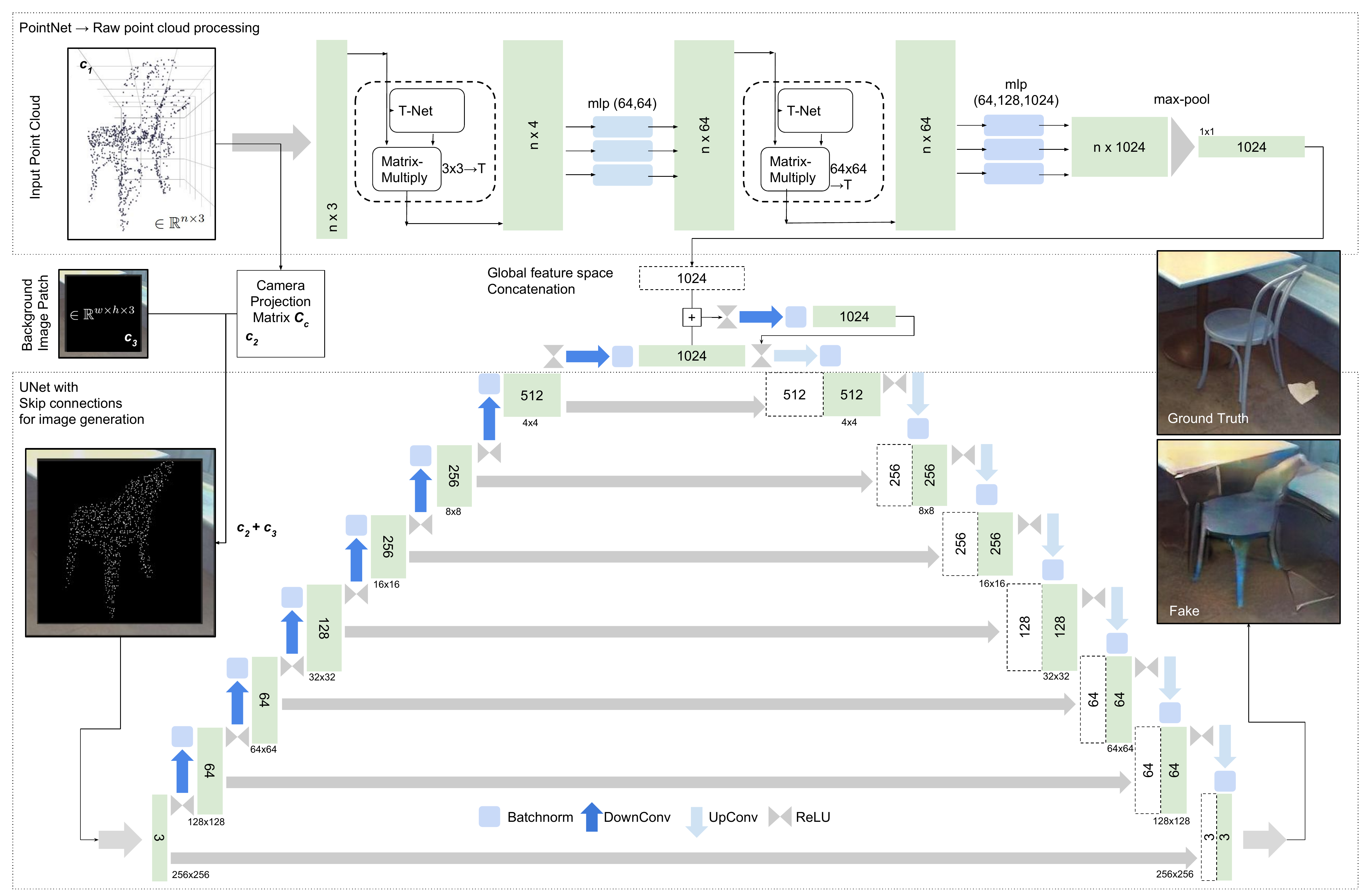}
\caption{Points2Pix Generator Architecture. The figure outlines the overall pipeline of the Points2Pix generator. In general we split the design into three areas: Top: the PointNet for a raw point-cloud processing; Bottom: Unet with skip connections for image generation; Middle: Global feature space concatenation from point-set (top) and image processing (bottom) pipeline. The model needs only a raw point-set as input, which acts as condition $c_1$. The point-sets coordinates will be directly processed by \textit{PointNet}. A projection into the image plane is used for UNet as input, whereas as the camera projection matrix $\mathbf{C_c}$ works as condition $c_2$. Additionally an arbitrary background patch $c_3$ is used for background generation.}
\label{fig:overview}
\end{figure*}

\subsubsection{Generative models}
Point Cloud GAN by \cite{DBLP:journals/corr/abs-1810-05795} is a famous approach for point-cloud generation. They do not perform any translation task, but they show that the common discriminator is not suitable for raw point-clouds. \cite{DBLP:journals/corr/AchlioptasDMG17} performs label to point-cloud translation by using representative learning and introduce several 3D GAN derivatives. A similar study is published by \cite{marrnet} with focus on latent space analysis. However, learning 3D representations to generate viewpoint based images is missing within the research community. Therefore, we propose our novel technique Points2Pix.

\section{Points2Pix}
We propose a novel cGAN architecture for generating photo-realistic images from pure point-clouds. Additionally, we describe conditions to bias the viewpoint, distance, shape and the background within the latent space. Therefore, we introduce the network architecture consisting of a generator $G$ (converting points to images), a discriminator $D$ and the specific loss. 

\subsection{Generator}
The objective of our generator $G(c_1,c_2,c_3)$ is to translate point-clouds into realistic-looking images, while using three conditions $c_1,c_2,c_3$. The whole architecture is shown in Figure \ref{fig:overview}. The design is inspired by \cite{DBLP:journals/corr/IsolaZZE16}, which serves as base. 
\subsubsection{Condition one}
First, $c_1$ as raw point-cloud $c_1=\{ x_1,...,x_n \}$ is processed by \textit{PointNet} \cite{qi2016pointnet}. The model samples $n = 1024$ points as input, applying an input transformation and aggregates global point features using fully connected layers and a generic max pooling (see equation \ref{equ:point}): 

\begin{equation}
\begin{aligned}
f(c_1) &\approx g(h(x_1), ..., h(x_n)) \;\;\;\ \;\;\;\ n=\{1 ... 1024\}
\end{aligned}
\end{equation}

However, in contrast to the basic \textit{PointNet} pipeline, the proposed model incorporates the the global 3D feature space ($n \times 1024$) using concatenation at the innermost part within the Image encoder-decoder (\textit{UNet}). Hence, $h(x_1), ..., h(x_n)$ are applied by the \textit{PointNet} part, $g$ is implicitly performed with the aid of \textit{UNets} decoder (see Fig. \ref{fig:overview}).

\subsubsection{Condition two}
The second condition denoted as $c_2 = \mathbf{C_c}$ is an image projection of the point-cloud using a perspective projection matrix $\mathbf{P}$
\begin{equation}
\mathbf{P} = \left[ \begin{array}{cccc}
s & 0 & 0 & 0 \\
0 & s & 0 & 0 \\
0 & 0 & -\frac{f_c}{f_c - n_c} & -1 \\
0 & 0 & -\frac{f_c \cdot n_c}{f_c - n_c} & 0 \\
\end{array}\right]
\; s = \frac{1}{\tan(\frac{fov}{2} \cdot \frac{\pi}{180})} \; \; \; \;
\mathbf{x}_{\textrm{pixel}} = \underbrace{\mathbf{P} \cdot \mathbf{T^{\textrm{cam}}_{\textrm{pc}}}}_{\mathbf{C_c}} \; x_i 
\end{equation}
\begin{equation}
\mathbf{y}_g(\mathbf{x}_{\textrm{pixel}}) = \frac{|\mathbf{x}_{\textrm{cam}}-x_i|}{d_{\textrm{max}}}  \; \; \; \;
\mathbf{y}_b(\mathbf{x}_{\textrm{pixel}}) = I(x_i)
\end{equation}
with a scaling $s$ according to the horizontal field of view denoted as $fov$ in degrees, near clipping plane denoted as $n_c$ and far clipping plane $f_c$. We encode radial depth ($\mathbf{y}_g(\mathbf{x}_{\textrm{pixel}}) \rightarrow$ green channel) with a normalized depth $d_{\textrm{max}}$ and intensities $I(x_i)$ of the measured reflectance for each point falling into the projection image ($\mathbf{y}_b(\mathbf{x}_{\textrm{pixel}}) \rightarrow$ blue channel). Before applying $P$, all points are transformed into the camera coordinate system using the extrinsic calibration $\mathbf{T}^{\textrm{cam}}_{\textrm{pc}}$. In this way we ensure the consistent viewpoint during training compared to the raw ground truth rgb image.

\subsubsection{Condition three}
Finally, the third condition $c3$ is an arbitrary image background patch constraining environmental texturing. A surrounding image patch of the object cropped from the data set centered at the object origin up to a size of $256 \times 256$ is extracted. During training the image background patch is compliant to the ground truth. In test-mode, background patches can be randomly mixed with point-clouds. 

Both, $c_2$  and $c_3$ are combined to an $256 \times 256$ input image, which is fed into a \textit{UNet} with skip connections. At the innermost part, down sampled input features are concatenated with the global 3D feature space from $c_1$. After up-sampling the output is a generated image with $256 \times 256$ pixels. Since, we use a cGAN for training, there is no need for an unstructured Loss. The assessment of the output is performed by the discriminator. As a note, we do not use a random noise vector $z$ (\ref{equ:noise}). Noise is only incorporated as dropout similar to \cite{DBLP:journals/corr/IsolaZZE16}.

\subsection{Discriminator}
We use the Markovian discriminator \textit{PatchGAN} \cite{DBLP:journals/corr/IsolaZZE16} that tries to distinguish between fake $D\left[G(c_1, c_2, c_3)\right]$ and real images $D[y]$ at the scale of $N \times N$ patches as well as possible. In contrast to \cite{DBLP:journals/corr/IsolaZZE16} we do not take the condition $c_1,c_2,c_3$ into account. The output depends only on the generated image. Therefore, it consists of an \textit{L1} term to force low-frequency correctness \cite{DBLP:journals/corr/ZhuPIE17} and is applied convolutionally across the image, averaging all responses. We only use $5$ convolutional layers with batch and instance normalization. In this way, it effectively solves the problem to be able to model high- and low frequency structures at once.

\begin{figure}[!t]
\centering
\includegraphics[width=0.7\textwidth]{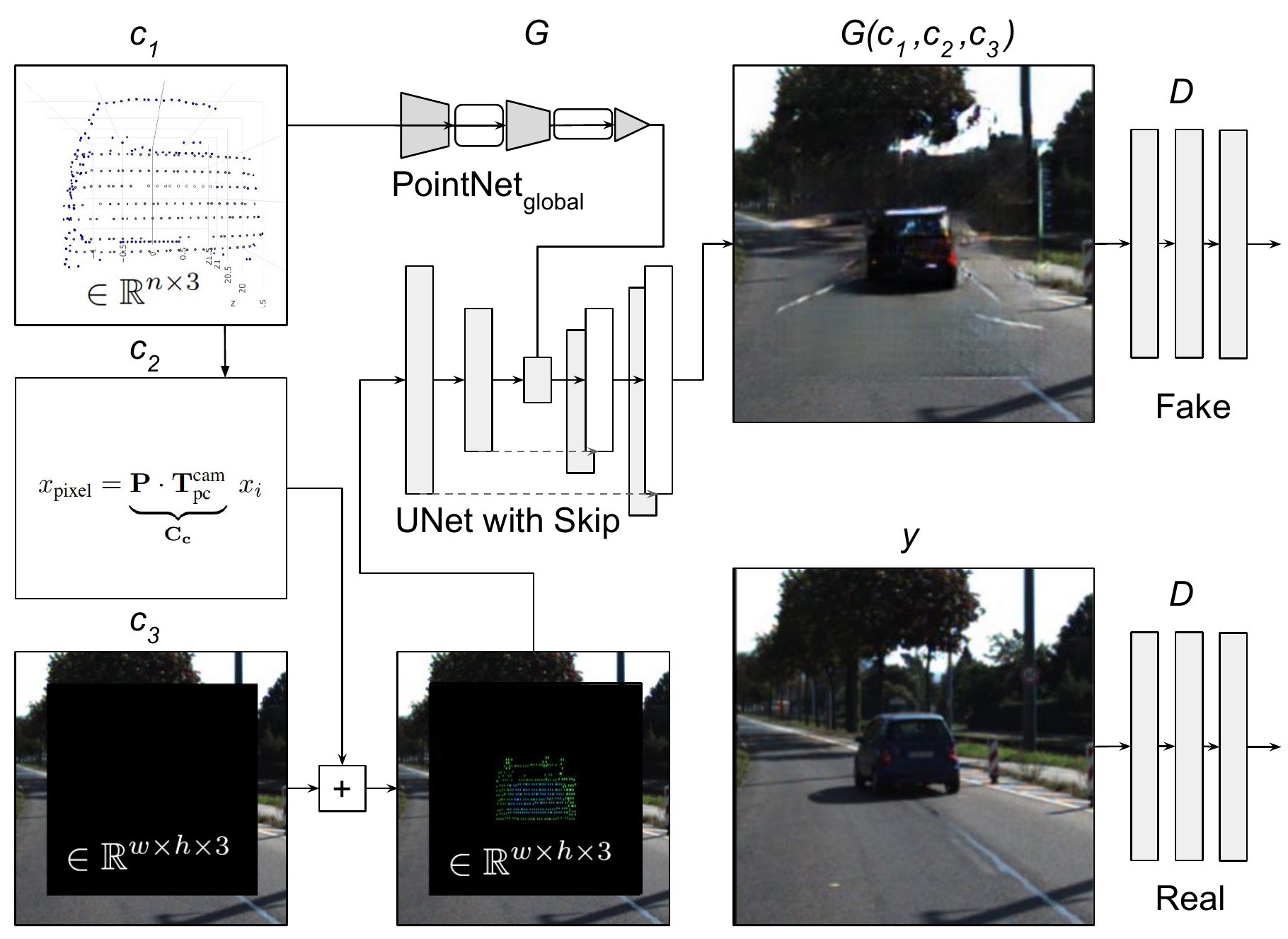}
\caption{Training Points2Pix: The figure outlines the competing training structure. The generators ($G$) output is a fake image based on its three conditions $c_1, c_2, c_3$. The discriminator $D$ has to distinguish between fake $D\left[G(c_1, c_2, c_3)\right]$ and real images $D[y]$.}
\label{fig:global}
\end{figure}

\subsection{Loss}
The objective of a basic GAN can be explained by an additive combination of the generative network $L_G$ loss and the discriminative network $L_D$ loss. In order to iteratively improve results during training $L_G$ should be reduced while $L_D$ grows ideally. Consequently the basic cGAN loss can be described as follows assuming the three input conditions ($c_1, c_2, c_3$):
\begin{equation}
\begin{aligned}
L_{cGAN}(G, D) = \mathbb{E}_y[\log(&D(c_2,c_3,y))] + \\ \mathbb{E}_{c_1,c_2,c_3}[\log(1-&D(c_2,c_3,G(c_1,c_2,c_3))]
\end{aligned}
\end{equation}

Random noise $z$ (\ref{equ:noise}) is only realized using dropout. Compared to the typical cGAN loss $L_{cGAN}$ (\ref{equ:loss}), the model does not involve all conditions into the discriminator. However, we implicitly force conditions to be compliant within the output by using a weighted $L_1$ term \cite{wang2018pix2pixHD} in the overall loss. This part describes the $L_1$ difference between the output and the ground truth. The final loss can be written as:

\begin{equation}
\begin{aligned}
L_{\textrm{Points2Pix}} &= L_{cGAN}(G, D) + \\&\lambda_{L1}\cdot
\mathbb{E}_{c_1, c_2, c_3, y}\left[\vert\vert y-G(c_1,c_2,c_3) \vert\vert _1\right]
\end{aligned}
\end{equation}

\section{Experiments}
We conduct experiments on KITTI \cite{Geiger:2012:WRA:2354409.2354978} for outdoor and SunRGBD \cite{Song2015} for indoor scenarios to explore the general validity of the method. Additionally, we show that the approach works for both, Lidar generated point-clouds and point-clouds coming from by RGB-D sensors. Following the recommendations of \cite{DBLP:journals/corr/IsolaZZE16}, the quality of the synthesized images is evaluated using an object based inception score. Furthermore, classification and diversity scores are added as additional assessment. Finally, we present some insights into our architecture decisions with additional ablation experiments.

\subsection{Metrics}
\label{sec_metrics}
To assess the realism of the produced images, YOLOv3 \cite{DBLP:journals/corr/abs-1804-02767} is used for validation. It is an off-the-shelf state of the art 2D object detector pre-trained on ImageNet and fine-tuned on the MS-Coco \cite{DBLP:journals/corr/LinMBHPRDZ14} data-set. This model includes overlapping classes in comparison to our experiments, e.g. cars (for KITTI) and chair (for SunRGBD). For the quantitative metrics we follow the instructions recommended by \cite{DBLP:journals/corr/WangG16}.

\subsubsection{Classification Score}
With the aid of YOLOv3 the number of correct detected classes is measured. This could be achieved due to object centered image patches in our experiments. The classification score $S_c$ ratio is then given by the detection ratio of fake images and ground truth ($TP \rightarrow$ true positives). The score could be directly affected by adjusting the confidence rate of the 2D object detector: $
S_c = TP_{\textrm{fake}}/TP_{real}$.

\subsubsection{Object based Inception Score}\footnote{We call it inception score, because its similar to the proposal of \cite{DBLP:journals/corr/SalimansGZCRC16}. We do not use an inception model.}
For positive results in terms of classification we measure the intersection over union $IoU_{\textrm{Points2Pix}}$ of the predicted bounding box $BB$ coming out of YOLOv3 for the ground truth and the accompanied fake image. 

\begin{equation}
IoU_\textrm{Points2Pix} = \frac{BB_{\textrm{fake}}\cap BB_\textrm{real}}{BB_{\textrm{fake}} \cup BB_{\textrm{real}}}\;\;\;\;| \;S_c=1
\end{equation}

\subsubsection{Diversity Score}
We measure the diversity ability of our cGAN to produce a wide spread of different output features using a diversity score. Our objective is to bias the shape, distance and 3D characteristics of the object. We collect randomly ten different background image patches, while keeping the point-cloud constant ($c1 = const$ and $c_2 = const$, $c_{3\rightarrow\{1...10\}}$). This leads to different output images that all should have the same 3D object inside. Therefore we compare the ground truth YOLOv3 results and all the fake images with the aid of calculating the mean $S_c$ and the mean $IoU$.

\subsection{Training Details}
We train the network on both data sets separately for $100$ epochs from scratch each, using the ADAM optimizer \cite{DBLP:journals/corr/KingmaB14}, with a learning rate of $0.0002$ and momentum parameters $\beta_1 = 0.5$, $\beta_2 = 0.999$ such as $\lambda_{L1} = 100$. For our background condition $c_3$, we use image patches with a border width of $15$ pixels. We found using objects containing at least 700 points in their point-cloud as a good trade-off for minimum point density as well as object size.

\textit{Kitti:} In a pre-processing step, we split the $7481$ training examples of the 3D object detection benchmark and use $3712$ samples for training and $3769$ for evaluation. Therefore, we generate more than $20k$ training images for the class \textit{car} only using $d_{\textrm{max}} = 60m$. Thus, each camera image is cropped centered at one labeled object with $256 \times 256$ pixels. At the same time, strongly occluded or truncated objects are skipped.

\textit{SunRGBD:} We extract 3267 images from the SunRGBD data-set containing the following classes: \textit{chair, table, desk, pillow, sofa} and \textit{garbage bin}. The split for training and validation is a 90/10 ratio. Image patches are extracted at the object center from the cameras point of view with a size of $256 \times 256$ pixels with $d_{\textrm{max}} = 4m$. The depth information comes from either MS kinect v1 or v2 and the Intel real-sense. Since, those sensors do not measure a reflectance, we only encode the radial depth inside the projection of $c_2$. Hence, the projection image contains one channel only. 

\begin{figure*}[!t]
\includegraphics[width=1\textwidth]{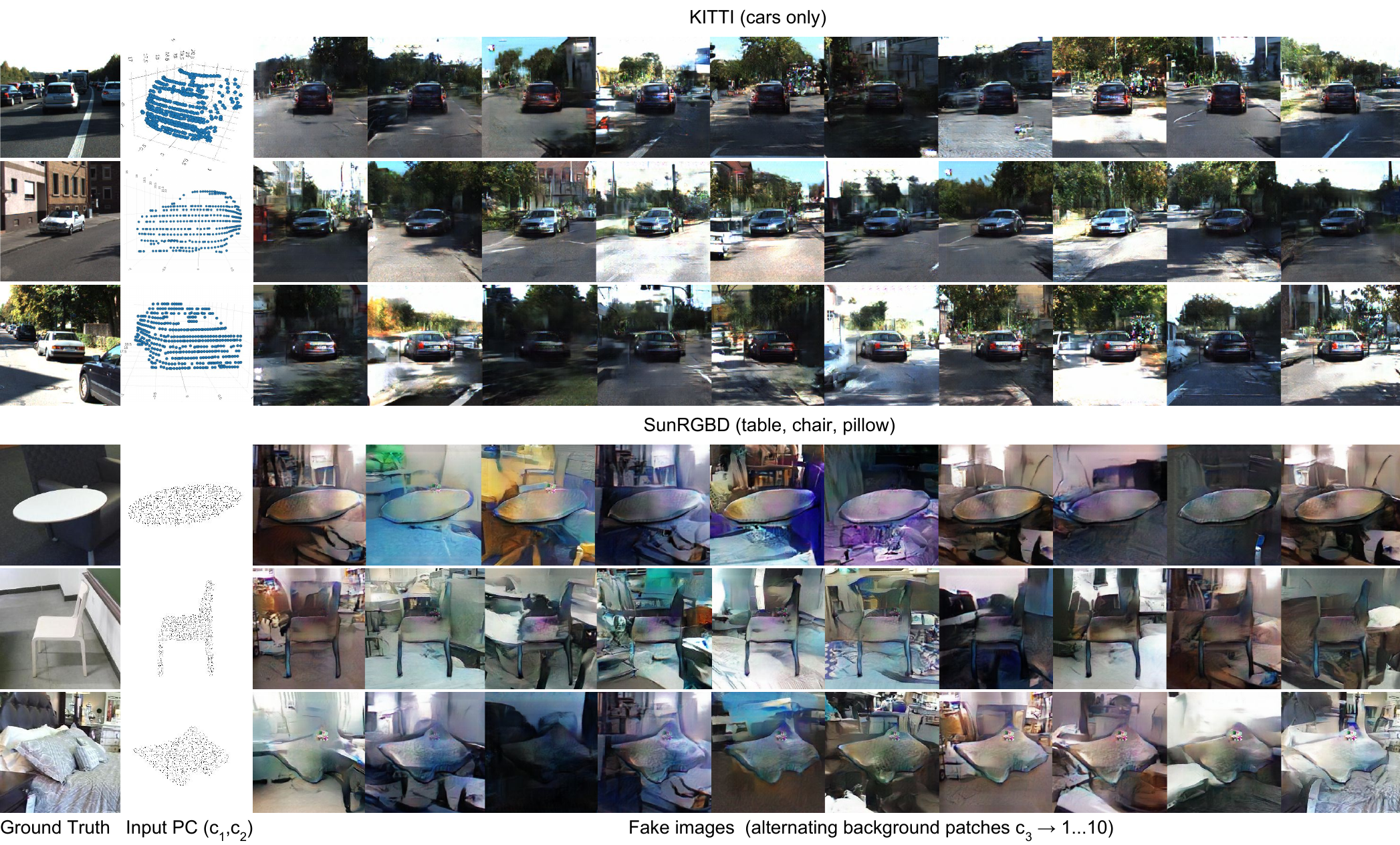}
\caption{Qualitative results of Points2Pix: The figure shows four different classes (cars (top) $\rightarrow$ 3 samples of KITTI; table, chair, pillow (bottom) $\rightarrow$ in each case one sample of SunRGBD). The results are taken from the test-set and never seen during training. The left column shows the ground truth image and the corresponding point-cloud in the second column. Fake images are generated based on a constant point-cloud $c_1, c_2 = const$ and ten alternating background patches $c_3$ (column 3-10). The model retains 3D characteristics of the objects.}
\label{fig:backgrounds}
\end{figure*}

\subsection{Results}
In Fig.~\ref{fig:backgrounds} we show qualitative results for both data-sets and four different classes. Widely distributed output images are produced by alternating the background while keeping the point-cloud constant. An interesting point is, that our model learns 3D characteristics. This could be proven with different outputs (backgrounds) where the objects geometry stays constant. Note, even the objects color stays the same apart from slight differences in reflections and illuminations. This means, the model associates a color with a specific 3D shape represented within the 3D latent feature space. Hence, alternating backgrounds do not affect the objects representation (geometry, color).

\begin{figure}[!t]
\centering
\includegraphics[width=0.75\textwidth]{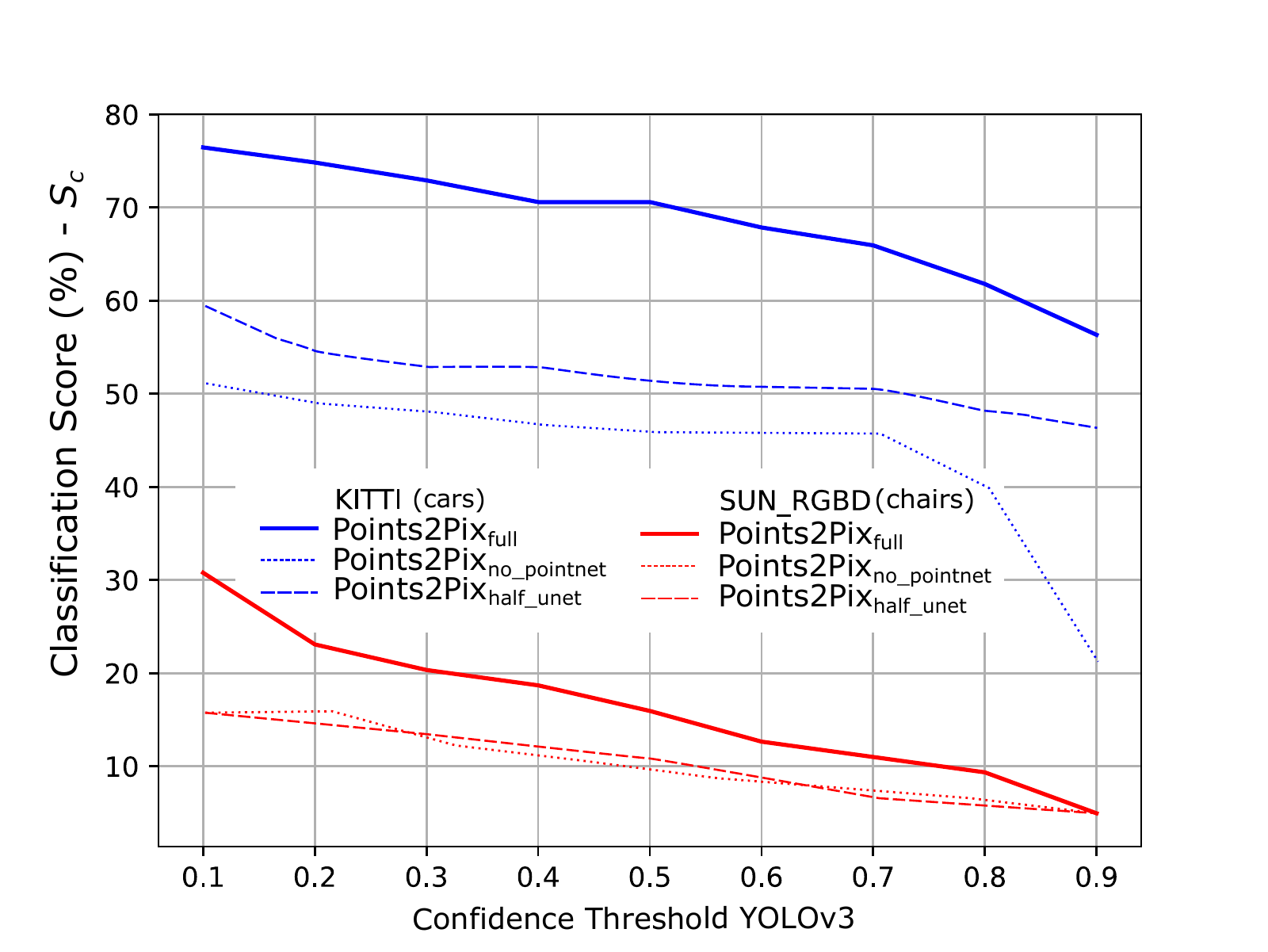}
\vskip -0.05in
\caption{Points2Pix classification score. The plot shows classification scores $S_c$ for KITTI and SUN-RGBD of our full Points2Pix architecture as well as two derivative architectures (see Fig.~\ref{fig:ablation_study}) over confidence thresholds used for object detection with YOLOv3. The full architecture outperforms for KITTI as well as for SunRGBD.}
\label{fig:plot}
\end{figure}

\begin{figure*}[!htb]
\centering
\includegraphics[width=0.95\textwidth]{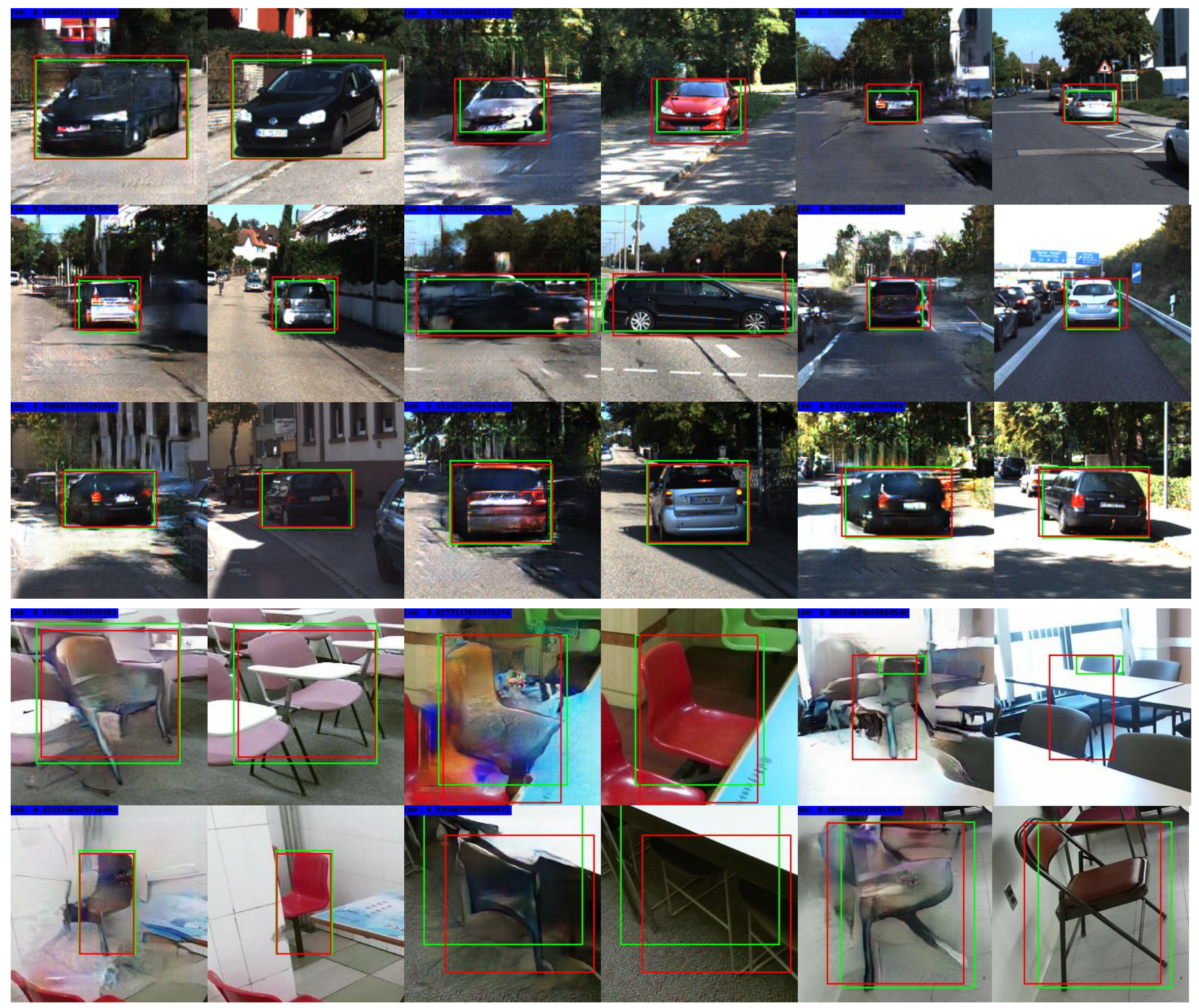}
\vskip -0.05in
\caption{Qualitative object based inception results. The figure shows several generated cars and chairs (left) together with their accompanied real images (right). Green bounding boxes indicate detections on the real rgb image patches and red boxes visualize the corresponding ones on the fake images. The blue value obtains the IoU of both.}
\label{fig:detection}
\end{figure*}

Tables~\ref{tab_yolo_score} and \ref{tab_diversity_score}, as well as Fig.~\ref{fig:plot} show quantitative results based on our metrics described in \ref{sec_metrics}. We achieve extreme positive results for KITTI ($S_c$, $IoU$) and sufficient values for SunRGBD. SunRGBD includes a higher number of occlusions which drastically affects the scores. Additionally, there are far less samples on each class compared to \textit{cars} in KITTI. Qualitative results of the inception score are shown in Fig.~\ref{fig:detection}. 

\begin{table}[!htb]
\begin{minipage}[t]{.48\linewidth}
\centering
\label{tab_yolo_score}
\resizebox{0.7\textwidth}{!}{
\begin{tabular}{|l|l|c|c|c|}
\hline
\multirow{2}{*}{\textbf{dataset}} & \multirow{2}{*}{\textbf{class}} & \multicolumn{3}{c|}{\textbf{$IoU_\textrm{Points2Pix}$}} \\
& & 0.3 & 0.5 & 0.7 \\
\hline
\hline
\textbf{KITTI} & car & 0.76 & 0.77 & 0.77 \\
\hline \hline
\multirow{2}{*}{} & \multirow{2}{*}{} & \multicolumn{3}{c|}{\textbf{$IoU_\textrm{Points2Pix}$}} \\
& & 0.1 & 0.2 & 0.3 \\
\hline \hline
\multirow{3}{*}{\textbf{SunRGBD}} & sofa & 0.52 & 0.77 & 0.77 \\
& table & 0.70 & - & - \\
& chair & 0.60 & 0.58 & 0.58 \\
\hline
\end{tabular}}
\vskip 0.15in
\caption{The object based inception score $IoU_\textrm{Points2Pix}$ is calculated on the test set for both data-sets. We show results for varying confidence thresholds, i.e. $0.3$, $0.5$, $0.7$ for KITTI and $0.1$, $0.2$, $0.3$ for SunRGBD.}
\end{minipage}%
\hfill
\begin{minipage}[t]{.48\linewidth}
\centering
\label{tab_diversity_score}
\resizebox{0.7\textwidth}{!}{
\begin{tabular}{|l|l|c|c|c|}
\hline
\multirow{2}{*}{\textbf{dataset}} & \multirow{2}{*}{\textbf{class}} & \multicolumn{3}{c|}{\textbf{$IoU_\textrm{Points2Pix}$}} \\
& & 0.3 & 0.5 & 0.7 \\
\hline
\hline
\textbf{KITTI} & car & 0.71 & 0.70 & 0.68 \\
\hline \hline
\multirow{2}{*}{} & \multirow{2}{*}{} & \multicolumn{3}{c|}{\textbf{$IoU_\textrm{Points2Pix}$}} \\
& & 0.1 & 0.2 & 0.3 \\
\hline \hline
\multirow{3}{*}{\textbf{SunRGBD}} & sofa & 0.16 & - &  - \\
& table & 0.24 & 0.22 & - \\
& chair & 0.45 & 0.37 & 0.33 \\
\hline
\end{tabular}
}
\vskip 0.15in
\caption{Diversity score is calculated on the test-set for both data-sets. Each sample is recomputed ten times with a random image background patches. A minus indicates no detections for the associated class.}
\end{minipage}%
\end{table}

\begin{figure}[!b]
\centering
\includegraphics[width=0.75\textwidth]{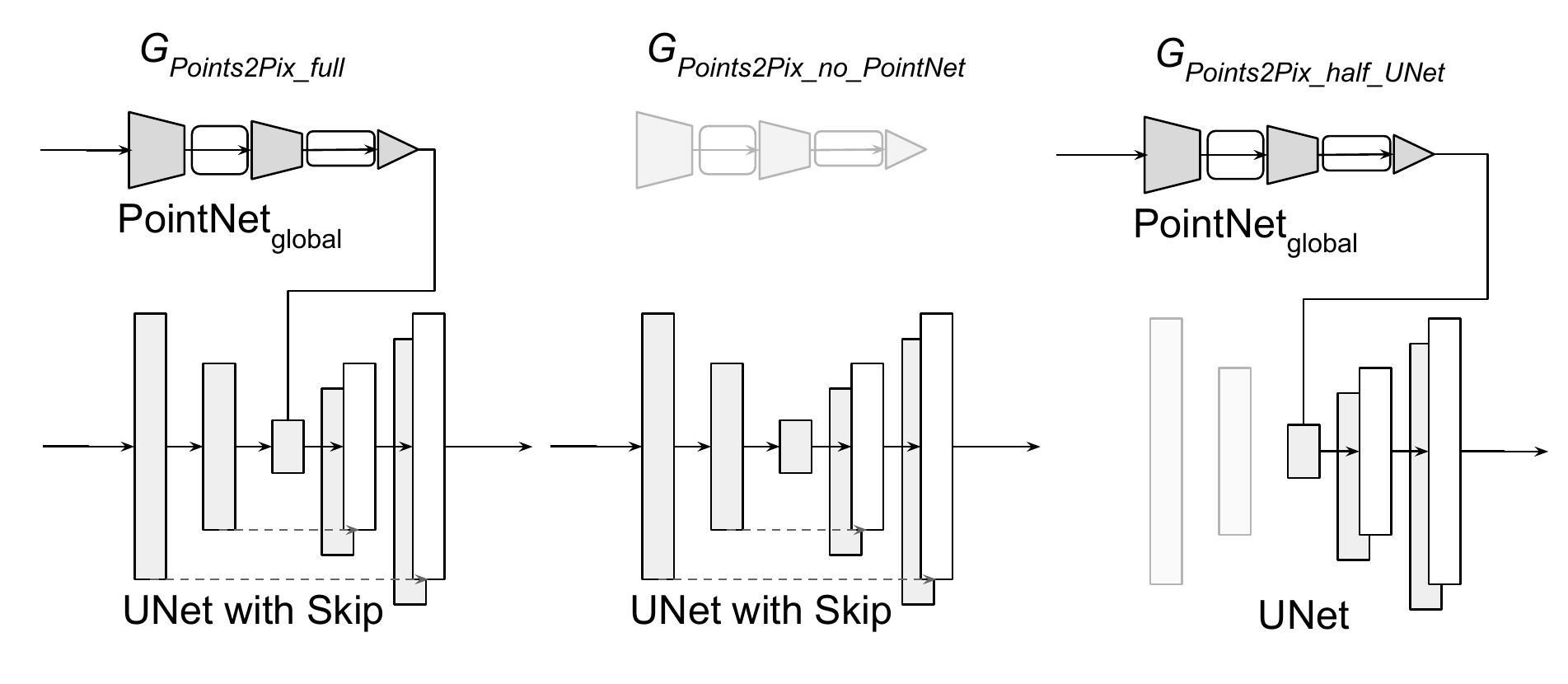}
\caption{Architectural review: Two derivatives from the basic Points2Pix $G(c_1, c_2, c_3)$ (left) generator are tested regarding their classification score (see Fig.\ref{fig:plot}). One on hand a \textit{Unet only} version $G(c_2,c_3)$ (middle), on the other hand a \textit{PointNet only} $G(c_1)$ (right) version is tested. The full model outperforms the others.}
\label{fig:ablation_study}
\end{figure}

\begin{figure}[!t]
\centering
\includegraphics[width=\textwidth]{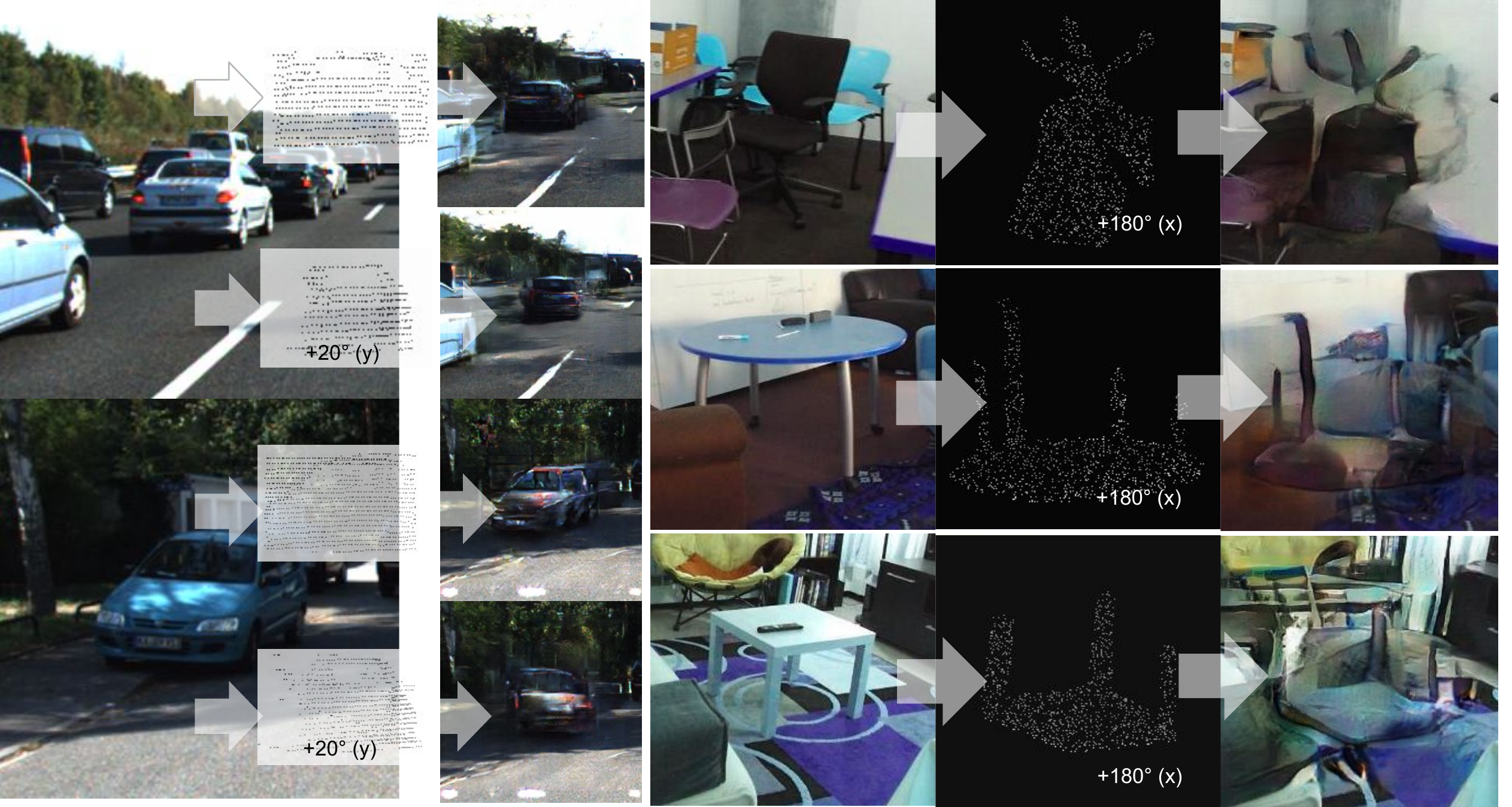}
\caption{Learning 3D representations: The full Points2Pix architecture learns 3D representations. The model offers a high flexibility in generation of different view points by adjusting condition $c_2$. The left part shows two examples of KITTI when rotating the point-cloud slightly by 20 degrees around the y-axis. The right (SunRGBD) shows the results when flipping the projection by 180 degrees around the x-axis.}
\label{fig:rotation}
\end{figure}

\subsubsection{Ablation study}

\paragraph{Architectural Review}
For completeness, we test two derivative architectures of our full pipeline (Fig.~\ref{fig:ablation_study}). In this way, we successfully show  a point-cloud to image translation only based on the point cloud itself (\textit{PointNet only}). Doing this, the whole training procedure runs much faster due to far less parameters to optimize. Nevertheless, sometimes a repeating noise with a high contrast similar to Moire effects appears, which indicates instabilities and uncertainties. Generated objects are in compliance with their 3D specifications, but in order to enlarge variance of the outputs and to control background conditions $c_2$ and $c_3$ are required. We found that the first part of the \textit{UNet} and the view-point dependent projection $c_2$ especially help to reduce the mentioned noise effects. They provide additional information in 2D space and stabilize the network. As a fallback we additionally test a \textit{Unet only} version  (Fig.~\ref{fig:ablation_study}). However, our full pipeline significantly outperforms the derivative architectures in terms of classification (Fig.~\ref{fig:plot}). 

\paragraph{Rotations}
To further emphasize the influence of $c_2$ and to show our models ability to constrain object view-points, we rotate all input points $x_i$ for $c_2$. We test that for KITTI with a rotation of 20 degrees around the y-axis and for SunRGBD with a rotation of 180 degrees around the x-axis (see Fig.~\ref{fig:rotation}). Note, that our point-cloud condition $c_1$ stays unmodified, because \textit{PointNet} approximates a symmetric function to be invariant of rotations. The test shows that rotations can be implicitly learned. This offers many opportunities in generating 3D data.

\section{Conclusion}
In this work, we propose a novel approach for 3D point-cloud to image translation based on conditional GANs. Our network handles multi-modal sources from different domains and is capable of the translating unordered point-clouds to regular image grids. We use three conditions to generate a high diversity, while being flexible and keeping 3D characteristics. We prove that the model learns 3D characteristics, what even makes it possible to sample images from different viewpoints. Those networks are applicable in a wide variety of applications, especially 3D texturing.

	\bibliographystyle{splncs04}
	\bibliography{egbib}

\end{document}